\documentclass[11pt]{article}
\usepackage[margin=0.75in]{geometry}
\usepackage[colorlinks,urlcolor=blue,linkcolor=blue,citecolor=blue]{hyperref}
\usepackage{color,array}
\usepackage{graphicx}
\usepackage{cite}
\usepackage{amsmath,amssymb,amsfonts}
\usepackage{algorithmic}
\usepackage{textcomp}
\usepackage{comment}

\begin{document}

\title{Agentic Artificial Intelligence (AI): Architectures, Taxonomies, and Evaluation of Large Language Model Agents}

\author{
Arunkumar V\\
University College of Engineering, Anna University\\
Tiruchirappalli, Tamil Nadu, India\\
\texttt{arunkumarv1530@gmail.com}
\and
Gangadharan G.R.\\
National Institute of Technology\\
Tiruchirappalli, India\\
\texttt{ganga@nitt.edu}
\and
Rajkumar Buyya\\
School of Computing and Information Systems\\
University of Melbourne, Australia\\
\texttt{rbuyya@unimelb.edu.au}
}
\date{}
 
\maketitle

\begin{abstract}
Artificial Intelligence is moving from models that only generate text to Agentic AI, where systems behave as autonomous entities that can perceive, reason, plan, and act. Large Language Models (LLMs) are no longer used only as passive knowledge engines but as cognitive controllers that combine memory, tool use, and feedback from their environment to pursue extended goals. This shift already supports the automation of complex workflows in software engineering, scientific discovery, and web navigation, yet the variety of emerging designs, from simple single loop agents to hierarchical multi agent systems, makes the landscape hard to navigate. In this paper, we investigate architectures and propose a unified taxonomy that breaks agents into Perception, Brain, Planning, Action, Tool Use, and Collaboration. We use this lens to describe the move from linear reasoning procedures to native inference time reasoning models, and the transition from fixed API calls to open standards like the Model Context Protocol (MCP) and Native Computer Use. We also group the environments in which these agents operate, including digital operating systems, embodied robotics, and other specialized domains, and we review current evaluation practices. Finally, we highlight open challenges, such as hallucination in action, infinite loops, and prompt injection, and outline future research directions toward more robust and reliable autonomous systems.

\vspace{1em}
\end{abstract}

\section*{Impact Statement}
Agentic AI changes the role of AI systems from conversation partners to active collaborators that can carry out tasks end to end. This paper investigates the architectures that let Large Language Models (LLMs) run complex workflows in software engineering, scientific discovery, and robotics, and explains how the field is moving from single agent loops to organized multi agent systems. We also call out key risks, including prompt injection and hallucination in action, and offer a practical roadmap for building autonomous systems that are robust, secure, efficient, and suitable for open ended real world environments.

\section{Introduction}

The field of Artificial Intelligence is moving from ``Generative AI'', which focuses on mapping inputs to static outputs, to Agentic AI, where systems are designed to actively change the state of their environment through perception, reasoning, and action. This shift is no longer driven only by prompt engineering. Frontier model families now expose stronger native reasoning behaviors, including configurable inference time reasoning budgets in dedicated reasoning models, which changes how planners and controllers are built in practice \cite{openai2024o1systemcard,snell2024testtimecompute}. At the same time, general purpose foundation models such as the OpenAI GPT 5 family, Google Gemini 3 Pro, and the Claude 4.5 family increasingly support structured tool use and multimodal interaction, yet in their base chat form they still operate turn by turn and do not maintain robust task state or permissions across long horizons \cite{openai2025gpt5systemcard,google2025gemini3,anthropic2025claude45systemcard}.

The idea of an autonomous agent has a long history in computer science. Classical work \cite{jennings2000agent} treated agents as encapsulated entities with situated autonomy, built on symbolic logic and hand written rules. Modern LLM based agents differ in important ways from both symbolic systems and reinforcement learning agents. Instead of depending on fixed symbolic representations or task specific policies learned through repeated interaction with an environment, contemporary agents draw on the probabilistic world knowledge encoded in foundation models and can generalize to unseen tasks with minimal task specific training, often achieving zero shot transfer \cite{piccialli2025agentai}. In practice, this allows agents to operate in open ended domains, from fixing bugs in large software repositories \cite{yang2024sweagent} to designing scientific workflows that combine hypothesis generation, code execution, and result interpretation \cite{ren2025scientific}.

The motivation for agentic systems is driven by workflows that cannot be completed within the context window, reliability envelope, or tool permissions of a single model call. However, the engineering response is not simply to make agents more autonomous. A major practical shift is toward controllable orchestration, where developers specify explicit state transitions and guardrails, and models fill in local decisions. This is visible in graph based orchestration frameworks and state machines that prioritize debuggability, checkpointing, and human approvals, often described as flow engineering \cite{langchain2024langgraph,openai2024swarm}. These orchestration choices interact directly with reliability and safety because the controller determines what actions are possible, when the agent can loop, and where verification and escalation occur.

Deploying Agentic AI also introduces risks that do not arise for static LLMs. Once a model can execute actions such as modifying files, running code, or operating a desktop interface, hallucinations can become concrete failures rather than incorrect text. Security threats also expand because agents must ingest untrusted content and act on it. Indirect prompt injection is a central example, where malicious instructions are embedded in web pages, documents, or tool outputs and then followed by an instruction obedient agent \cite{liu2024formalizing}. These risks are amplified by new interaction modes such as native computer use, where agents operate user interfaces through screenshots and mouse and keyboard actions \cite{anthropic2024computerusetool,openai2025operator}. In parallel, tool integration is becoming more standardized at the infrastructure layer. The Model Context Protocol provides a common way to expose tools and resources to agents, reducing fragmentation in connector schemas and enabling governance patterns such as allowlists and audit logging at the protocol boundary \cite{anthropic2024mcp,anthropic2024mcpspec}.

\subsection*{Contributions of this paper}

Existing surveys on LLM based agents and agentic AI have mainly taken four complementary perspectives:
(i) broad overviews of LLM based autonomous agents and their applications \cite{wang2024survey,xi2025rise},
(ii) conceptual and definition oriented treatments of agentic AI that go beyond LLMs \cite{abou2025agentic},
(iii) methodology focused surveys centered on tool use, planning, and feedback learning \cite{li2025review,qu2025tool},
and (iv) recent reviews that describe LLM agents mainly from a methodology point of view \cite{luo2025large}.
In contrast, this survey is explicitly architecture and engineering focused: we start from a formal POMDP based agentic control loop and organize the literature around how concrete systems are built, deployed, and evaluated in practice, including inference time reasoning, controllable orchestration, and standardized tool connectivity.

\noindent Concretely, our main contributions are:

\begin{itemize}
    \item \textbf{Unified architecture focused taxonomy.}
    We propose a unified taxonomy that decomposes LLM based agents into six modular dimensions:
    Core Components (perception, memory, action, profiling), Cognitive Architecture (planning, reflection),
    Learning, Multi Agent Systems, Environments, and Evaluation.
    This architecture first view connects each component to the underlying control loop and complements prior surveys
    that group work mainly by applications \cite{wang2024survey,xi2025rise} or by paradigms such as tool use, planning, and feedback learning \cite{li2025review}.

    \item \textbf{Engineering and systems perspective.}
    Beyond high level methodology, we highlight concrete design choices that matter in deployed systems:
    memory backends and retention policies,
    agent computer interfaces and computer use actions,
    the shift from JSON style function calling to code as action,
    standardized connector layers such as MCP,
    and orchestration controllers that enforce typed state and explicit transitions.
    Compared to methodology centered surveys \cite{luo2025large},
    our focus is on how these pieces are assembled into robust, monitorable agent systems.

    \item \textbf{From autonomous loops to controllable graphs.}
    We give a unified treatment of multi agent systems that links classical MAS ideas with modern LLM based frameworks.
    We describe interaction patterns such as chain, star, mesh, and explicit workflow graphs, and we analyze frameworks including
    CAMEL, AutoGen, MetaGPT, LangGraph, Swarm, and MAKER under the same lens.
    This extends earlier multi agent surveys that often discuss collaboration mechanisms in isolation \cite{tran2025multiagent,zhu2025multiagentbench}.

    \item \textbf{Holistic view of environments, evaluation, and safety.}
    We place evaluation directly in the architectural space using the CLASSic dimensions
    of cost, latency, accuracy, security, and stability.
    We link architectural choices such as hierarchical planning, code execution, graph controllers, and computer use actions to concrete failure modes including hallucination in action, infinite loops, and prompt injection.
    We also summarize lessons from enterprise oriented benchmarks and deployment studies \cite{khamis2025agentic,liu2024agentbench,wornow2025classic}.
\end{itemize}

Among existing works, Wang \emph{et al.} \cite{wang2024survey} and Xi \emph{et al.} \cite{xi2025rise} provide broad overviews of LLM based autonomous agents but devote comparatively less attention to detailed engineering of perception modules, memory backends, connector standards, and controllable orchestration. Luo \emph{et al.} \cite{luo2025large} survey the field through a methodology centered taxonomy focused on planning, tool use, learning, and collaboration.
Our taxonomy is complementary: we derive an explicit systems architecture that covers perception, memory, the agent brain, and action, and we use this to guide how to build robust agents rather than only cataloging what existing agents can do.

The rest of the paper is organized as follows. Section II provides formal definitions and background. Section III presents the taxonomy of the Agentic AI ecosystem. Section IV details the unified architecture of Agentic AI. Section V analyzes Multi Agent Systems and their interaction topologies. Section VI discusses Environments and Applications where the agents operate. Section VII addresses Evaluation and Safety assessment. Section VIII outlines Challenges and Future Directions of the Agentic AI ecosystem, followed by concluding remarks in Section IX.

\section{Background and Definitions}

This section outlines the evolution of the agent concept, formalizes the definition of an agent using a mathematical decision-making framework, and presents a holistic taxonomy that constitute a modern agentic system. 

\subsection{From Symbolic and RL Agents to LLM based Agents}

The concept of an ``agent'' has changed substantially over the history of computer science. Early work focused on symbolic agents, as described in foundational studies such as \cite{jennings2000agent}, which relied on predefined rules, formal logic, and fixed constraints to operate in closed environments. These systems were often robust within their design envelope but brittle when confronted with situations that fell outside their programmed rules. As research progressed, attention shifted toward Reinforcement Learning (RL) agents, where policies are learned through repeated interaction with an environment and through explicit trial and error. These agents are very effective on specific control tasks, yet they usually lack the generalization ability needed for open ended reasoning or flexible natural language interaction and they demand large amounts of data and careful sample efficiency for each new task \cite{busoniu2008marl}.

The modern LLM based agent can be seen as a third paradigm. Instead of depending on hard coded rules or narrowly defined reward functions, these agents use a pretrained Large Language Model as a general purpose cognitive controller. In this view, the LLM acts as a reasoning engine that can be augmented with external memory and execution modules \cite{mialon2023augmented}, and the design focus shifts from training policies to prompt design, tool integration, and orchestration. This allows agents to transfer the semantic knowledge learned from large internet corpora into concrete action oriented tasks \cite{li2025review}. In practice, such agents can often generalize in a zero shot manner to new environments that include both software development and robotics, achieving behaviors that were not feasible with purely symbolic systems or with standard RL agents alone.

\subsection{Formal Definition: The Agentic Control Loop}

Although concrete architectures differ, there is growing agreement on how to describe the basic mathematical structure of an LLM based agent. Rather than viewing an agent only as a policy, it is useful to treat it as a dynamic control system that operates within a Partially Observable Markov Decision Process (POMDP). We write the agent system $\mathcal{A}$ as a tuple
\[
\mathcal{A} = \langle \mathcal{S}, \mathcal{O}, \mathcal{M}, \mathcal{T}, \pi \rangle ,
\]
where $\mathcal{S}$ is the state space, $\mathcal{O}$ the observation space, $\mathcal{M}$ the internal memory space, $\mathcal{T}$ the action or tool space, and $\pi$ the policy.

At each discrete time step $t$, the system moves through a cycle of linked functions.

\subsubsection{Perception Function \texorpdfstring{($\Phi$)}{(Phi)}}
The agent does not have access to the full environment state $S_t \in \mathcal{S}$. Instead, it receives a partial observation $O_t$ through a perception function $\Phi$, which may include multimodal encoders (for example vision models such as CLIP) or text based wrappers:
\begin{equation}
O_t = \Phi(S_t) \quad \text{with} \quad O_t \subset S_t .
\end{equation}

\subsubsection{Memory Update Mechanism \texorpdfstring{($\mu$)}{(Mu)}}
In contrast to stateless reinforcement learning agents, LLM based agents maintain a mutable internal state $M_t$. This state is updated by a function $\mu$ that combines the new observation $O_t$, the previous reasoning trace $Z_{t-1}$, and the execution feedback $E_{t-1}$:
\begin{equation}
M_t = \mu(M_{t-1}, O_t, Z_{t-1}, E_{t-1}) .
\end{equation}
This update step subsumes retrieval augmented generation (RAG) \cite{mialon2023augmented}, where $\mu$ selects and injects relevant information from the agent's long term memory into the current context.

\subsubsection{Cognitive Planning \texorpdfstring{($\Psi$)}{(Psi)}}
A defining feature of Agentic AI is the latent reasoning step $Z_t$ (the thought or plan) that is produced before any external action is taken. We model this as a probabilistic inference process parameterized by the LLM $\theta$:
\begin{equation}
Z_t \sim P_\theta(Z_t \mid M_t, O_t) .
\end{equation}
The variable $Z_t$ can represent a simple chain of thought or a more structured hierarchical plan. In advanced architectures such as RAP \cite{hao2023rap}, this planning phase is implemented as a recursive tree search over possible future trajectories.

\subsubsection{Action Policy \texorpdfstring{($\pi$)}{(Pi)} and Execution}
Finally, the agent chooses and executes an action $A_t$ from the tool space $\mathcal{T}$. The policy $\pi$ is conditioned directly on the reasoning trace, so that actions are explicitly grounded in the internal plan:
\begin{equation}
A_t \sim \pi_\theta(A_t \mid Z_t, M_t) .
\end{equation}
The environment then reacts to this action through a state transition $S_{t+1} \leftarrow \text{Env}(S_t, A_t)$ and produces feedback $E_t$, which flows back into the next perception and memory update steps, closing the control loop \cite{li2410agent}.

\section{Taxonomy}

\subsection{Holistic Taxonomy of the Agentic AI Ecosystem}
\label{subsec:historic_taxonomy}

\begin{figure*}[t]
\centering
\includegraphics[width=\textwidth]{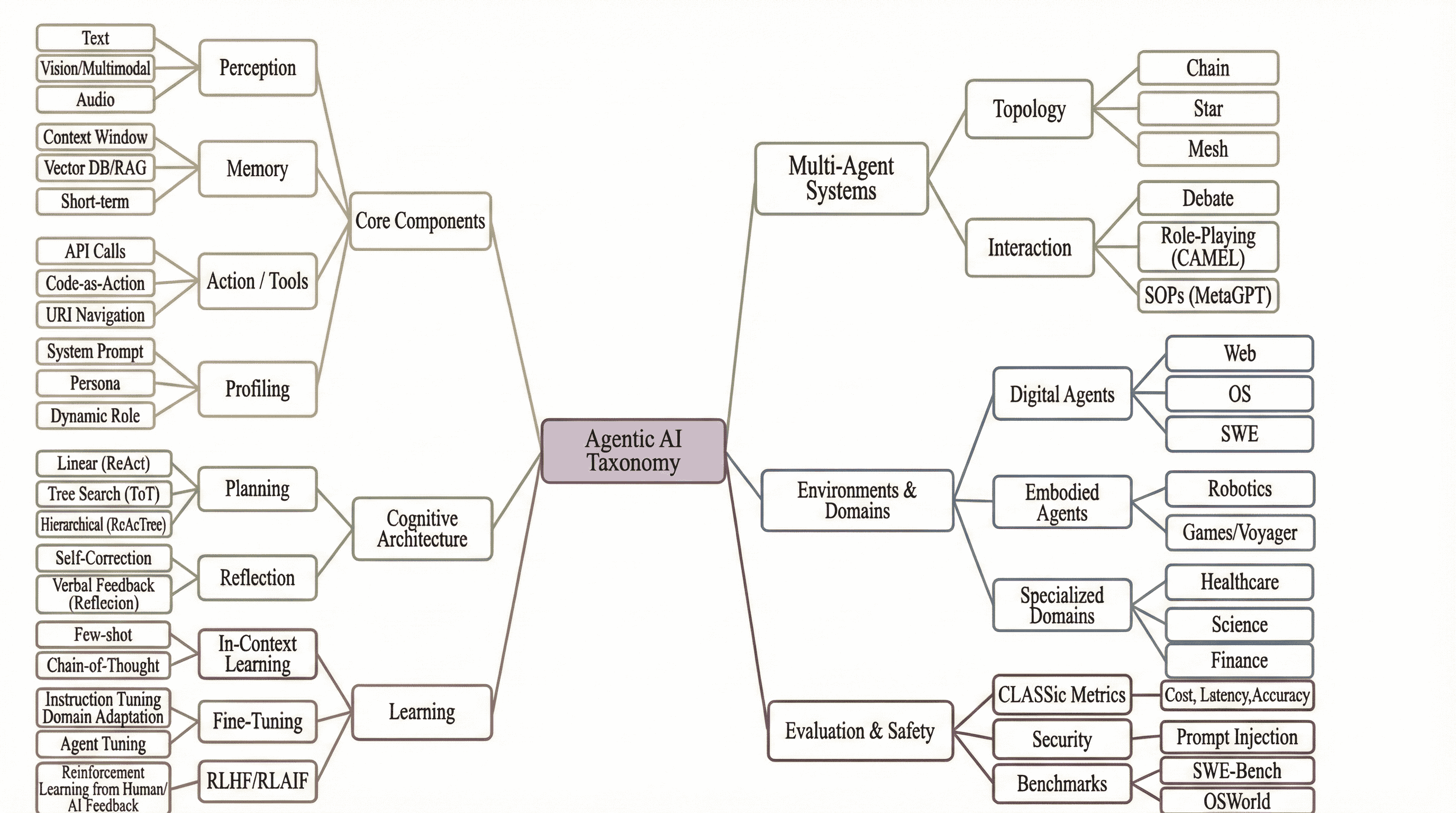}
\caption{\textbf{Taxonomy of the Agentic AI ecosystem.} The figure organizes the literature into six main dimensions: Core Components, Cognitive Architecture, Learning, Multi Agent Systems, Environments, and Evaluation. Together, these dimensions trace the field's progression from simple text based loops to complex hierarchical systems that can operate in open ended environments.}
\label{fig:taxonomy}
\end{figure*}

Fig.~\ref{fig:taxonomy} gives a high level view of the agentic AI ecosystem and the way its architectures have evolved over time. The field has grown by expanding along six connected dimensions, moving from individual capabilities inside a single model to systems that coordinate many agents and include explicit mechanisms for evaluation and safety.

\paragraph{Core Components}
The foundation of any agent is its interface with the world. Perception has moved from processing only raw text to handling visual, multimodal, and audio inputs, which allows agents to interpret complex graphical user interfaces. Memory has shifted from short lived context windows to persistent storage backed by vector databases and retrieval augmented generation (RAG), which supports continuity over long horizons. In the same spirit, action interfaces have developed from fixed API calls into more flexible Code as Action and direct URI navigation. Profiling completes this layer by defining the agent's identity through system prompts and dynamic roles, so that its behavior remains consistent across different tasks.

\paragraph{Cognitive Architecture}
The cognitive architecture dimension describes how agents reason. Early systems relied on linear planning loops such as ReAct. To deal with more complex problems, recent work has adopted hierarchical structures that use tree search methods, for example Tree of Thoughts, and recursive decomposition as in ReAcTree. To improve reliability, these planners are complemented by reflection mechanisms, including self correction and verbal feedback methods such as Reflexion, which allow agents to critique and refine their plans before they act.

\paragraph{Learning}
The learning dimension captures how agents acquire and improve capabilities over time. At one end of the spectrum are in context methods, such as few shot prompting and Chain of Thought, which are temporary and live entirely in the prompt. At the other end are permanent weight updates through fine tuning, including instruction tuning and agent specific tuning. On top of these, alignment techniques such as RLHF and RLAIF adjust agent behavior using feedback from humans or from other models, and are increasingly used to shape decision making in realistic settings.

\paragraph{Multi Agent Systems}
As tasks grow beyond the capacity of a single model, the taxonomy extends to multi agent systems. This dimension distinguishes between interaction styles that range from adversarial debate to cooperative role playing, as in CAMEL, and structured workflows based on standard operating procedures, as in MetaGPT. It also highlights different communication topologies, where agents can be organized in chains for sequential processing, in star shaped configurations with a central coordinator, or in mesh like swarms for more decentralized collaboration.

\paragraph{Environments and Domains}
Agents are also defined by the environments in which they operate. The taxonomy groups these into digital agents that work inside web browsers, operating systems, or software engineering tools, embodied agents in robotics and games such as Voyager, and specialized domains including healthcare, science, and finance. Each class of environment comes with its own constraints and affordances, which in turn impose specific requirements on perception, memory, and action design.

\paragraph{Evaluation and Safety}
The final dimension addresses how agentic systems are assessed and secured. Evaluation frameworks have moved beyond single accuracy scores and now incorporate the CLASSic metrics of cost, latency, accuracy, security, and stability. Security research focuses in particular on defending against threats such as prompt injection. At the same time, standardized benchmarks such as SWE-Bench for software engineering and OSWorld for operating system control provide shared references for tracking progress across different architectures.

We use this taxonomy as the organizing structure for the remainder of the paper. It allows us to analyze each dimension in depth, from single agent cognitive architectures to multi agent coordination patterns and the evaluation and safety practices that are required for real world deployment.

\section{The Unified Architecture of Agentic AI}

\begin{figure*}[t]
\centering
\includegraphics[width=\textwidth]{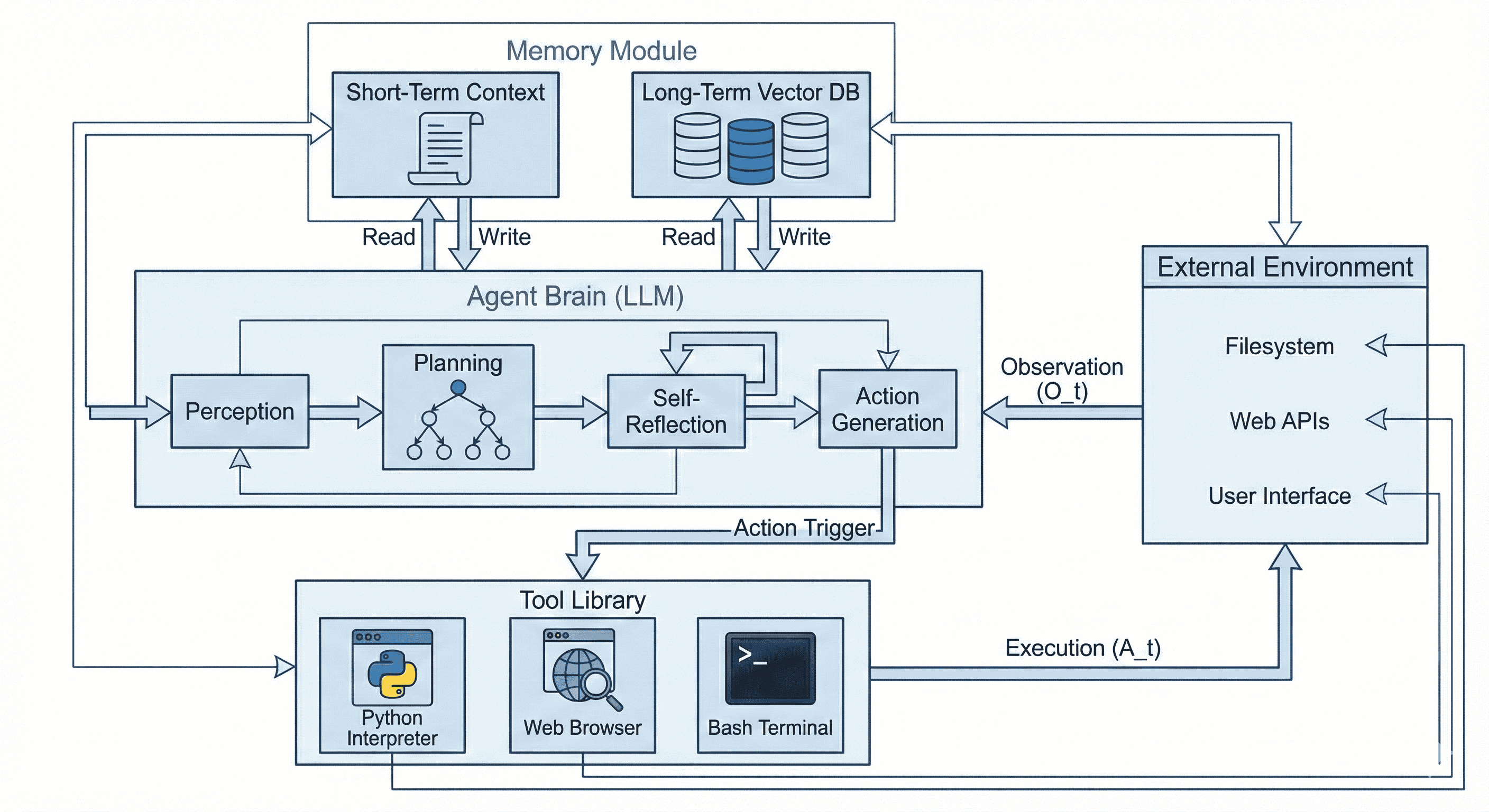}
\caption{\textbf{The unified architecture of Agentic AI.} The system is shown as a modified POMDP loop. The agent brain at the center transforms each observation ($O_t$) into a reasoning trace ($Z_t$) using hierarchical planning and self reflection. A dual stream memory module at the top supports context retrieval, while a tool library at the bottom executes code based actions ($A_t$) that change the external environment on the right.}
\label{fig:architecture}
\end{figure*}

We view the architecture of an autonomous agent as a cognitive pipeline that turns perception into action through a central decision making process. As illustrated in the feedback loop in Fig.~\ref{fig:architecture}, this system integrates modular resources with a reasoning engine that runs over time. In line with the taxonomy in Fig.~\ref{fig:taxonomy}, we break this pipeline into three layers: i) the Core Components, which provide interfaces for perception, memory, action, and profiling; ii) the Cognitive Architecture, which carries out planning and reflection; and iii) Learning, which describes how agents acquire and refine their capabilities.

\subsection{Core Components}

Modern agent architectures are built from modular components coordinated by a cognitive process. Building on infrastructure layers \cite{kumar2024building} and construction taxonomies \cite{luo2025large}, we write the core components of an agent $A$ as a tuple
\[
A = \langle \Phi, M, T, P \rangle ,
\]
where $\Phi$ is perception, $M$ is memory, $T$ is the set of actions and tools, and $P$ is profiling.

\subsubsection{Perception}

Perception ($\Phi$) is the interface between the agent and its environment. Early agents such as ReAct operated entirely in the text domain. In contrast, recent systems make use of multimodal large language models (MLLMs) that ground their reasoning in higher dimensional sensory inputs.

\begin{table*}[t]
\centering
\caption{\textbf{Evolution of agent perception modules.} The table summarizes how agents ground themselves in digital and physical environments and highlights the move from pure text processing to multimodal vision and audio encodings.}
\label{tab:perception_modules}
\resizebox{\textwidth}{!}{%
\begin{tabular}{|l|l|l|l|l|}
\hline
\textbf{Model / Framework} & \textbf{Input modality} & \textbf{Grounding mechanism} & \textbf{Target environment} & \textbf{Critical challenge} \\ \hline
\textbf{WebVoyager} \cite{he2024webvoyager} & Vision (screenshots) & CLIP based visual encoder & Open web browsers & \textit{Visual clutter}: distracted by ads and pop ups; high token cost for images. \\ \hline
\textbf{AppAgent} \cite{chi2024appagent} & Vision + touch coordinates & Visual to action mapping & Smartphone apps (Android) & \textit{Dynamic UIs}: struggles with animations or fading elements. \\ \hline
\textbf{SeeAct} \cite{zheng2024seeact} & Hybrid (HTML + vision) & Cross attention & Web forms & \textit{Grounding gap}: hallucinates $(x,y)$ coordinates for small buttons. \\ \hline
\textbf{Magma} \cite{yang2025magma} & Video + proprioception & Set of Mark (SoM) & Robotics & \textit{Latency}: real time video processing is too slow for reactive control. \\ \hline
\textbf{AudioGPT} \cite{huang2024audiogpt} & Audio and speech & Audio foundation models & Voice assistants & \textit{Turn taking}: difficulty handling interruptions in live conversation. \\ \hline
\textbf{3D LLM} \cite{hong20243dllm} & 3D point clouds & Spatial encoders & Embodied navigation & \textit{Spatial resolution}: low fidelity in detecting small obstacles. \\ \hline
\end{tabular}%
}
\end{table*}

\paragraph{Multimodal grounding for digital and embodied settings}
Modern perception modules move beyond text-only inputs to support UI-level grounding (screenshots, touch coordinates, and hybrid DOM+vision) and embodied sensing (video, audio, and 3D geometry). WebVoyager and AppAgent demonstrate screenshot-driven and coordinate-based control in browsers and mobile apps, but both expose a persistent grounding bottleneck: mapping high-level intent to precise UI targets, where small elements and dynamic layouts cause drift or hallucinated clicks \cite{he2024webvoyager,chi2024appagent,zheng2024seeact}. For time-varying scenes, Magma extends perception to continuous video streams for robotic manipulation \cite{yang2025magma}. Beyond vision, AudioGPT integrates speech/audio understanding for voice-first interaction \cite{huang2024audiogpt}, while 3D-LLM injects point-cloud representations to preserve spatial affordances that are lost in 2D projections \cite{hong20243dllm}. Table~\ref{tab:perception_modules} summarizes representative designs, grounding mechanisms, and dominant failure modes.

\subsubsection{Memory}

To behave as persistent agents rather than one off sessions, systems need mechanisms that preserve state over time and support consistent behavior. Memory components address these needs and go beyond simple vector retrieval to include structure, retention policies, and explicit mechanisms for summarization and deletion.

\begin{table*}[t]
\centering
\caption{\textbf{Evolution of agentic memory architectures.} We compare approaches for long term state persistence. The retention strategy column indicates how each system prevents uncontrolled growth of context, for example by forgetting, summarizing, or paging.}
\label{tab:memory_architectures}
\resizebox{\textwidth}{!}{%
\begin{tabular}{|l|l|l|l|l|}
\hline
\textbf{Architecture} & \textbf{Memory structure} & \textbf{Retrieval strategy} & \textbf{Retention strategy} & \textbf{Primary benefit} \\ \hline
\textbf{Generative Agents} \cite{park2023generative} & Natural language stream & Scoring by recency and relevance & Reflection and summarization & Social consistency and long horizon coherence in simulations \\ \hline
\textbf{MemoryBank} \cite{zhong2024memorybank} & Hierarchical clusters & Hierarchical traversal & Exponential decay & Keeps memory relevant through forgetting \\ \hline
\textbf{ChatDB} \cite{hu2024chatdb} & Symbolic SQL tables & SQL queries & Exact storage & High precision for structured and numerical state \\ \hline
\textbf{MemGPT} \cite{xu2023memgpt} & Paged long term memory & Controller driven retrieval & Explicit paging and summarization & Manages long contexts through externalized memory control \\ \hline
\textbf{MemInsight} \cite{zhao2024meminsight} & Insight level summaries & Semantic clustering & Merge and compress & Converts episodic traces into compact semantic memories \\ \hline
\textbf{MemAgent} \cite{ma2025memagent} & Active read and write store & Policy learned retrieval & Policy driven pruning & Learns what to store, summarize, and discard across sessions \\ \hline
\end{tabular}%
}
\end{table*}

\paragraph{Memory as retrieval, structure, and retention policy}
A central systems decision is how an agent manages working context under cost and attention constraints: long-context prompting scales poorly with irrelevant history, so retrieval-augmented generation remains important for focusing on task-relevant state \cite{mialon2023augmented}. MemoryBank shows that hierarchical summaries can outperform raw-log prompting by making retrieval structured and self-pruning \cite{zhong2024memorybank}, while Chain-of-Agents distributes extremely long contexts across coordinated workers when a single prompt is insufficient \cite{ge2025chain}. For long-horizon coherence, Generative Agents maintain a timestamped memory stream \cite{park2023generative}, and ACAN-style alignment improves recall beyond vanilla similarity retrieval \cite{hong2025memory}. Retention policies (summarize/forget/prune) prevent unbounded growth: MemInsight compresses episodic traces into semantic “insights” \cite{zhao2024meminsight}, and MemAgent treats memory management as a learned decision process across sessions \cite{ma2025memagent}. When state is naturally structured, ChatDB provides high-precision symbolic memory via SQL \cite{hu2024chatdb}, while MemGPT formalizes paging between short context and external stores under an explicit controller \cite{xu2023memgpt}. Table~\ref{tab:memory_architectures} compares these designs by structure, retrieval, and retention strategy.

\subsubsection{Action and Tools}

While the cognitive architecture (discussed below) provides the reasoning blueprint, the action component allows the agent to actually change its environment. This module turns the semantic intent produced by the planning layer into concrete operations. Over time, the space of possible actions has moved from fixed, predefined function calls toward more flexible code execution and interface level navigation. For clarity, we group these mechanisms into four main paradigms: API calls, Code as Action, Agent Computer Interfaces, and embodied Vision Language Action (VLA).

\begin{table*}[t]
\centering
\caption{\textbf{Evolution of agentic action spaces.} The table contrasts the main ways agents execute tasks. Recent production systems add computer use actions for generic GUI control, alongside code execution and constrained tool calls.}
\label{tab:tools}
\resizebox{\textwidth}{!}{%
\begin{tabular}{|l|l|l|l|l|}
\hline
\textbf{Paradigm} & \textbf{Action space} & \textbf{Representative frameworks} & \textbf{Key advantage} & \textbf{Primary limitation} \\ \hline
\textbf{API based} & Predefined structured calls & Toolformer \cite{schick2023toolformer}, Gorilla \cite{patil2023gorilla} & Safety through restricted scope & Statelessness and tool schema friction \\ \hline
\textbf{Code as action} & Executable scripts (often Python) & CodeAct \cite{wang2024codeact}, Voyager \cite{wang2024voyager} & Rich control flow and state & Sandboxing risk and runtime errors \\ \hline
\textbf{ACI based} & Curated shell or IDE actions & SWE agent \cite{yang2024sweagent} & Efficient context use & Requires interface design and upkeep \\ \hline
\textbf{Computer use actions} & Mouse, keyboard, screenshots & Claude computer use \cite{anthropic2024computeruse}, Operator \cite{openai2025operator} & Works on arbitrary GUIs without app APIs & Latency, brittleness, and injection risk \\ \hline
\textbf{Embodied VLA} & Continuous motor primitives & Gemini Robotics \cite{deepmind2025gemini} & Physical grounding & Latency and sim to real gaps \\ \hline
\end{tabular}%
}
\end{table*}

\paragraph{From constrained APIs to code, interfaces, and embodied control}
Agent action spaces have expanded from constrained API/function calls to executable programs, UI-level control, and continuous embodied policies. API-centric systems such as Toolformer and Gorilla emphasize safety and tool selection through structured calls, but often require brittle schema prompting and ad hoc state passing across steps \cite{schick2023toolformer,patil2023gorilla}. At the infrastructure layer, MCP standardizes tool discovery and invocation so controllers can connect to evolving tool catalogs with governance boundaries such as allowlists, authentication, and audit logging \cite{anthropic2024mcp,anthropic2024mcpspec}.

In parallel, “code as action” uses executable scripts as the control interface—enabling variables and control flow and aligning better with code-heavy pretraining distributions—illustrated by CodeAct and Voyager’s reusable skill library \cite{wang2024codeact,wang2024voyager}. For long-horizon digital work, curated Agent Computer Interfaces (e.g., SWE-agent) reduce context load by exposing simplified, agent-friendly shells/IDEs \cite{yang2024sweagent}. Finally, computer-use actions (screenshots + mouse/keyboard) enable generic GUI operation but introduce latency and broader injection risk \cite{anthropic2024computeruse,openai2025operator}, while embodied VLA systems push toward direct perception-to-motor control in robotics \cite{deepmind2025gemini}. Table~\ref{tab:tools} summarizes these paradigms and their tradeoffs.

\subsubsection{Profiling}

Memory gives an agent continuity of experience, while profiling gives it a stable character. This module specifies the identity, role, and implicit constraints of the agent \cite{cheng2024exploring}. In practice, profiling corresponds to the system prompt or persona that shapes behavior, for example, ``You are a senior Python engineer who writes clear and well tested code.'' Such profiles narrow the search space and improve alignment. An agent whose profile emphasizes security will naturally favor safer plans than one that prioritizes speed. More advanced designs allow for dynamic roles, so that an agent can switch, for instance, from a writer role to an editor role as a task moves from drafting to revision.

\subsection{Cognitive Architecture}

The cognitive architecture is the agent’s decision core: it decomposes goals into action sequences $A_t$ and monitors execution for inconsistency and failure. Unlike classical symbolic planners with hand-written rules and explicit environment models, LLM-based agents leverage probabilistic world knowledge to propose plans and revise them under feedback. Following our taxonomy, we describe this layer through \emph{planning} (how trajectories are constructed) and \emph{reflection} (how trajectories are evaluated and improved).

\subsubsection{Planning}

A foundational agentic pattern is ReAct \cite{yao2023react}, which interleaves intermediate reasoning with environment interaction, in contrast to purely latent Chain-of-Thought prompting \cite{wei2022chain}. ReAct executes trajectories
$\tau=\{o_1,z_1,a_1,o_2,z_2,a_2,\dots\}$,
where rationales $z_t$ condition actions $a_t$ and subsequent observations, improving groundedness but remaining vulnerable to myopia and error propagation (e.g., early mistakes leading to unproductive loops).

To reduce myopia, many approaches add explicit branching and search. Tree of Thoughts treats intermediate thoughts as search nodes and explores alternatives via breadth/depth strategies \cite{yao2024tree}. LATS further connects agent planning to MCTS by using the model as both proposal and evaluator, enabling selection among candidates before committing to irreversible tool calls \cite{zhou2024lats}.

A recent shift is that frontier reasoning models internalize parts of search at inference time. OpenAI’s o1 line emphasizes improved performance with increased test-time ``thinking'' \cite{openai2024learning2reason,openai2024o1systemcard}, aligning with work on test-time compute scaling \cite{snell2024testtimecompute} and subsequent releases such as o3 \cite{openai2025o3systemcard}. This does not remove external planning modules, but changes their role: modern systems increasingly combine controllable inference-time reasoning budgets with external controllers that enforce safety, state persistence, and tool permissions.

As task complexity grows, flat planners still face context and modularity limits, motivating hierarchical decomposition. ReAcTree distributes long-horizon planning across recursive sub-agents with explicit control flow \cite{choi2025reactree}, while GoalAct uses a two-tier structure with a global milestone planner and local executors \cite{chen2025goalact}. These designs improve interpretability and fault isolation but can increase token and latency costs in large or noisy environments.

\begin{table*}[t]
\centering
\caption{\textbf{Comparative analysis of agentic cognitive architectures.} We compare planning methodologies by topology, search strategy, and inference cost. \textit{Token complexity} estimates inference overhead relative to a standard zero shot prompt ($N$: steps, $b$: branching factor, $d$: depth, $B$: inference time compute budget).}
\label{tab:cognitive_arch}
\resizebox{\textwidth}{!}{%
\begin{tabular}{|l|l|l|l|l|l|}
\hline
\textbf{Method} & \textbf{Core mechanism} & \textbf{Reasoning topology} & \textbf{Token complexity} & \textbf{Key advantage} & \textbf{Critical limitation} \\ \hline
\textbf{Standard CoT} \cite{wei2022chain} & Latent reasoning steps & Linear chain & Low ($\sim 1\times$) & Low latency & Error propagation without recovery \\ \hline
\textbf{ReAct} \cite{yao2023react} & Interleaved thought and action & Linear loop & Medium ($N\times$) & Groundedness & Myopic behavior and looping \\ \hline
\textbf{Reflexion} \cite{shinn2023reflexion} & Verbal reinforcement buffers & Cyclical loop & High ($k \cdot N\times$) & Self correction & Hallucinated critiques and context overflow \\ \hline
\textbf{Tree of Thoughts} \cite{yao2024tree} & External branching search & Tree structure & Very high ($b^d\times$) & Backtracking and global search & Latency and combinatorial growth \\ \hline
\textbf{LATS} \cite{zhou2024lats} & MCTS with value feedback & Graph or tree & Very high & Strong accuracy on hard tasks & Depends on evaluators and heavy compute \\ \hline
\textbf{Reasoning models (o1, o3)} \cite{openai2024learning2reason,openai2024o1systemcard,openai2025o3systemcard} & Internalized inference time search & Adaptive implicit tree & Variable ($\propto B$) & Compute quality tradeoff inside the model & Higher cost and limited transparency \\ \hline
\textbf{ReAcTree} \cite{choi2025reactree} & Recursive sub agent spawning & Hierarchical tree & Medium high & Modular long horizon solving & State synchronization complexity \\ \hline
\end{tabular}%
}
\end{table*}

To improve efficiency without additional labels, test-time optimization methods refine behavior using signals extracted from agent trajectories (e.g., RISE and test-time self-improvement) \cite{qu2024rise,acikgoz2025ttsi}. Proactive architectures further separate internal deliberation from user-facing outputs to bound interactive latency \cite{liu2024proactive}. Table~\ref{tab:cognitive_arch} positions these approaches alongside prompted search, hierarchy, and inference-time compute scaling.

\subsubsection{Reflection}

Reflection mechanisms allow agents to critique and adapt based on their own trajectories, transforming sparse success/failure signals into actionable guidance. A common pattern is \emph{verbal reinforcement}: Reflexion stores natural-language critiques of failures and conditions future attempts on these lessons rather than updating weights \cite{shinn2023reflexion}. Complementarily, \emph{self-correction protocols} apply critique before committing to outputs. Self-Refine implements an iterative generate--critic--revise loop \cite{madaan2023selfrefine}, while CRITIC reduces purely internal confirmation loops by requiring tool-interactive validation (e.g., interpreters or search) before accepting revisions \cite{gou2024critic}.

Reflection is especially important under tool failures (timeouts, schema mismatches, repeated retries). PALADIN trains recovery behaviors on large collections of failure trajectories, enabling diagnosis and corrected retries that reduce loop-like failures \cite{vuddanti2025paladin}. Expel extracts reusable rules from past mistakes (e.g., safety or rollback heuristics) and applies them to new tasks, supporting both episode-level repair and cross-episode generalization \cite{zhao2023expel}.

\subsection{Learning Paradigms}

The learning dimension describes how agent capability improves over time, spanning ephemeral in-context adaptation, permanent weight updates, and non-parametric accumulation of executable skills. Early agents relied mainly on in-context learning (examples and heuristics in prompts), which is easy to deploy but short-lived and increasingly costly as prompts grow. This motivates agent tuning, where useful behaviors are internalized in weights via trajectory data: Agent-FLAN and FireAct fine-tune models on agent rollouts, and FireAct shows that ``hot'' trial-and-error trajectories can outperform prompt-only methods by shifting task logic from long prompts into compact parameters and reducing inference cost \cite{chen2024agent,chen2024fireact}. However, effective tuning must avoid overfitting to surface formats rather than robust reasoning patterns.

Beyond supervised tuning, scalable oversight is a bottleneck. RLAIF uses AI feedback to generate preference labels \cite{lee2024rlaif}, enabling process-level reward models such as AgentRM and AgentPRM to provide dense step-wise guidance without heavy human annotation \cite{xia2025agentrm,xi2025agentprm}. These signals support test-time search and selection among candidate trajectories, complementing hand-designed heuristics. Self-improvement pipelines further show that agents can bootstrap performance by mining failures and generating synthetic corrections, reporting sizable gains on web-agent tasks \cite{patel2024large}. Finally, agents can improve without weight updates by storing and reusing executable skills: Voyager maintains an external skill library of successful code fragments that can be retrieved and composed, accelerating open-ended learning while avoiding catastrophic forgetting \cite{wang2024voyager}.

\section{From Single to Multi Agent Systems}

\begin{figure*}[t]
\centering
\includegraphics[width=\textwidth]{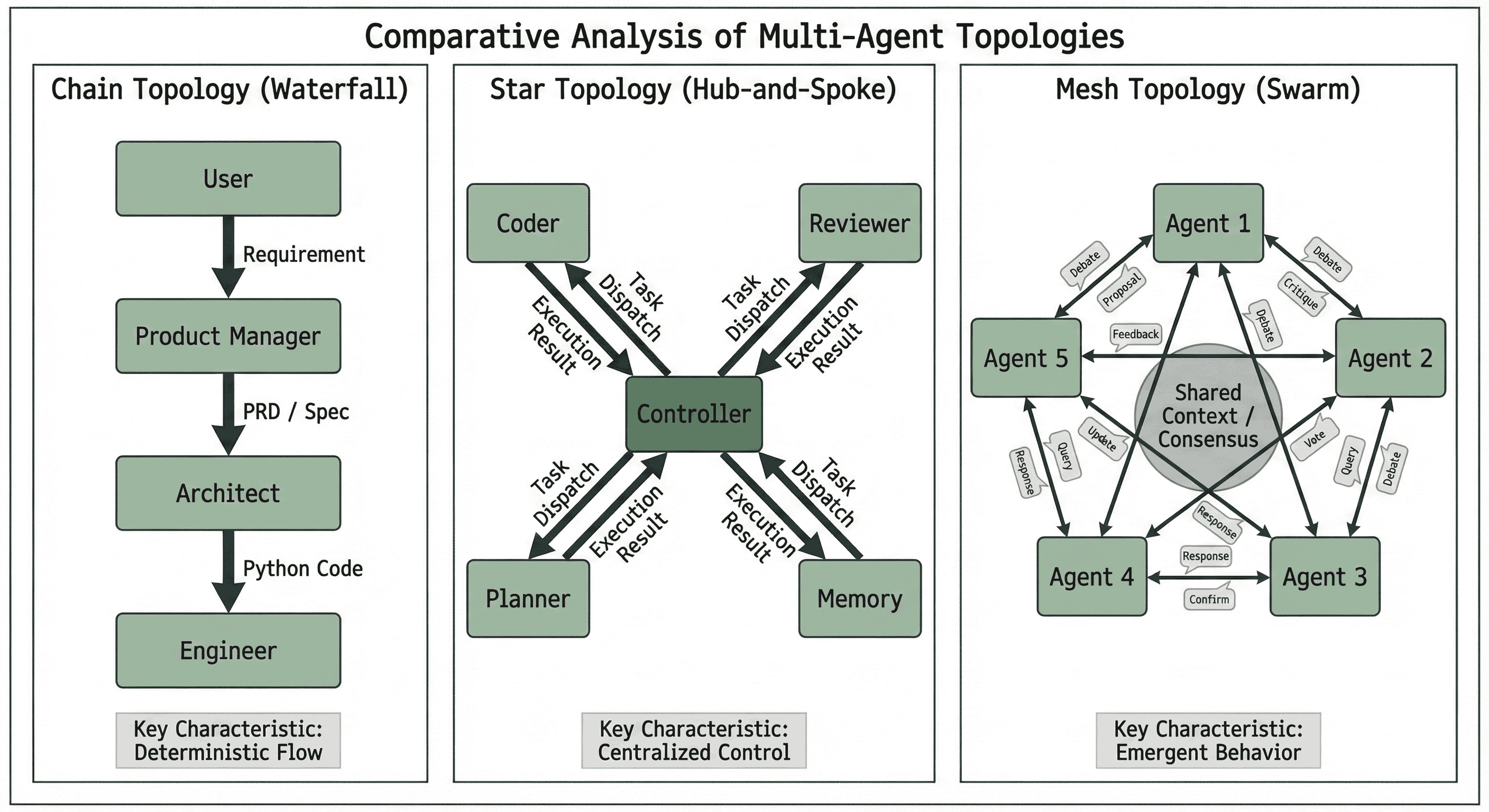}
\caption{\textbf{Communication Topologies in Multi-Agent Systems.} We classify collaboration patterns into three dominant structures: (Left) Chain Topology, utilized by MetaGPT to enforce Standard Operating Procedures (SOPs) via sequential hand-offs; (Center) Star Topology, employed by AutoGen where a Controller agent dispatches tasks to specialized workers; and (Right) Mesh Topology, used in social simulations like Generative Agents to enable dynamic, unstructured interaction.}
\label{fig:mas_topology}
\end{figure*}

While single agent architectures have shown strong capabilities in isolation, many real world problems still exceed the context window, domain expertise, or cognitive capacity of a single model. This has led to a shift toward multi agent systems (MAS), in which groups of specialized agents collaborate to complete workflows. As illustrated in Fig.~\ref{fig:mas_topology}, the effectiveness of a multi agent system depends heavily on its interaction topology. Following \cite{tran2025multiagent}, these structures are usually grouped into three main patterns: i) chain or waterfall, where tasks are passed along a fixed sequence of agents for rigid workflows; ii) star or hub and spoke, where a central controller coordinates specialized workers; and iii) mesh or swarm, where agents interact in a more decentralized and dynamic way, for example during brainstorming.

\begin{table*}[t]
\centering
\caption{\textbf{Analysis of multi agent collaboration frameworks.} We map prominent frameworks to the coordination structures in Fig.~\ref{fig:mas_topology} and include graph based orchestration as an increasingly common production pattern.}
\label{tab:mas_frameworks}
\resizebox{\textwidth}{!}{%
\begin{tabular}{|l|l|l|l|l|l|}
\hline
\textbf{Framework} & \textbf{Topology} & \textbf{Role structure} & \textbf{Communication style} & \textbf{Conflict resolution} & \textbf{Ideal use case} \\ \hline
\textbf{CAMEL} \cite{li2023camel} & 1 to 1 mesh & Symmetric & Inception prompting & None (chat loop) & Ideation and brainstorming \\ \hline
\textbf{AutoGen} \cite{wang2024autogen} & Star or mesh & Dynamic & Message passing & Human in the loop & Prototyping heterogeneous tools \\ \hline
\textbf{MetaGPT} \cite{hong2024metagpt} & Chain (waterfall) & Static roles & SOP documents & Sequential handoff & Complex software engineering \\ \hline
\textbf{ChatDev} \cite{qian2024chatdev} & Chain & Static phases & Waterfall handoffs & Reviewer rejection & End to end app generation \\ \hline
\textbf{DyLAN} \cite{liu2024dylan} & Dynamic graph & Dynamic selection & LLM optimized routing & Importance scoring & Focused reasoning collaboration \\ \hline
\textbf{LangGraph} \cite{langchain2024langgraph} & Workflow graph & Developer defined & State machine execution & Guard nodes and approvals & Production flow engineering \\ \hline
\textbf{Swarm} \cite{openai2024swarm} & Star with handoffs & Lightweight specialists & Handoff routines & Controller selection & Controllable coordination patterns \\ \hline
\textbf{TradingAgents} \cite{xiao2024tradingagents} & Mesh (market) & Heterogeneous & Auction or debate & Market equilibrium & Economic simulation \\ \hline
\textbf{MAKER} \cite{cognizant2025maker} & Hierarchical & Supervisor and worker & Cross examination & Verifier agent & High reliability logic tasks \\ \hline
\end{tabular}%
}
\end{table*}

\subsection{Graph based orchestration and flow engineering}

A key industry shift is the move from open ended multi agent chat loops toward explicit workflow graphs. Rather than relying on a generic manager agent to decide what to do next in free form dialogue, many production systems model the workflow as a state machine where nodes represent tool calls or LLM invocations and edges represent permissible transitions. This approach is often described as flow engineering, since the developer designs the control structure and the agent fills in local decisions within that structure.

LangGraph is a representative framework that operationalizes this idea by treating agent execution as graph traversal with explicit state persistence, checkpoints, and controlled cycles \cite{langchain2024langgraph}. This makes long horizon behavior more debuggable and easier to align with organizational constraints, because developers can insert guard nodes, approval steps, and typed state updates at specific points in the graph.

OpenAI Swarm provides a complementary view that emphasizes lightweight agent handoffs and routines for orchestrating specialist behaviors, and it is positioned as a reference implementation for controllable coordination patterns \cite{openai2024swarm}. In practice, these patterns align with a broader trend: orchestration is increasingly specified as an explicit controller layer, while the LLM focuses on local reasoning, tool parameterization, and recovery. Graph based designs also interact naturally with safety and evaluation, since the graph boundary defines what actions are even possible, which can reduce the frequency and severity of runaway loops.

\subsection{Architectures of Collaboration and Role Playing}

The core mechanism that drives many multi agent systems is role playing. The CAMEL framework \cite{li2023camel} was an early and influential example. It introduced inception prompting to set up autonomous cooperative dialogues by assigning specific personas (for instance a stock trader) and starting a role flip conversation. CAMEL showed that agents could guide one another through multi step tasks using only these role prompts. However, CAMEL uses a simple one to one mesh topology that can drift into unproductive loops because there is no central decision maker. AutoGen \cite{wang2024autogen} generalizes this idea by treating agents as conversable entities that can be wired together in more flexible ways. Developers can specify custom interaction graphs, and many practical setups adopt a star topology in which a user proxy or manager agent delegates subtasks to tools and worker agents and then aggregates their outputs. This structure also supports human in the loop oversight, which is critical for safety sensitive applications. Moving beyond fixed graphs, DyLAN \cite{liu2024dylan} proposes a dynamic agent network. Instead of using a static interaction pattern, DyLAN estimates the contribution of each agent at every step and selects collaborators based on an importance score. Agents that are not helpful for the current problem are muted, which reduces token usage and cost, while relevant agents remain active. This adaptive routing leads to more efficient and focused collaboration during complex reasoning chains.

\subsection{Organizational Metaphors: The Chain Topology}

A distinct sub-class of multi-agent systems draws inspiration from human corporate structures, implementing Standard Operating Procedures (SOPs) to streamline collaboration.

\subsubsection{Software Swarms}
MetaGPT \cite{hong2024metagpt} formalized this approach by encoding SOPs directly into agent prompts. By assigning specific roles such as Product Manager, Architect, and Engineer, MetaGPT forces agents to generate standardized deliverables (e.g., PRDs, API Designs) that serve as strict inputs for the next agent in the waterfall. This effectively functions as an ``operating system'' for agents, reducing the hallucination rate by transforming unstructured chat into a rigorous workflow. This concept was further exemplified in ChatDev \cite{qian2024chatdev}, which simulates an entire software company where agents self-organize into design, coding, and testing phases, achieving a 30\% reduction in bugs compared to single-agent coding.

\subsubsection{Hierarchical Verification}
Linear handoffs can propagate errors if an upstream agent fails. To address this, MAKER \cite{cognizant2025maker} advances the organizational metaphor by introducing a ``cross-examination'' phase. MAKER demonstrates that by decomposing tasks into granular steps and using distinct ``Verifier'' agents to challenge the output of ``Worker'' agents, systems can execute million-step reasoning chains with near-zero error accumulation. This aligns with findings from BOLAA \cite{liu2024bolaa}, which proved that orchestrating multiple smaller, specialized agents often outperforms a single massive model by distributing the cognitive load. This hierarchical safety extends to supervisory architectures. OVON \cite{gosmar2025hallucination} and The Good Parenting framework \cite{kwartler2024parenting} propose ``Parenting'' topologies, where a ``Reviewer'' agent holds a distinct system prompt focused solely on critique and safety guidelines. Empirical results show that filtering ``Child'' agent outputs through a Supervisor can reduce hallucination rates by up to 100\% in controlled environments.

\subsection{Social Simulation and Debate: The Mesh Topology}

While chains optimize for efficiency, Mesh Topologies optimize for creativity and diversity. In these systems, interaction is decentralized, allowing emergent behaviors to arise from agent-to-agent dynamics.

\subsubsection{Adversarial Debate}
Dialectical interaction has been proven to enhance reasoning performance. Research on Multiagent Debate \cite{du2024debate} revealed that allowing multiple LLM instances to propose conflicting answers and critique each other's reasoning leads to a convergence on truth. This ``society of minds'' approach is further refined in Adaptive Debate \cite{zhang2025adaptive}, where agents assume specialized adversarial roles (e.g., ``Devil's Advocate'') to stress-test solutions. Recent work extends this to Communication Games \cite{neurips2024chemical}, where agents play referential games to align on descriptions of complex molecules. This constrained communication forces agents to develop precise, compositional language, improving generalization to novel structures.
 
\subsubsection{Economic and Social Simulation}

Mesh topologies are also used to model complex systems. TradingAgents \cite{xiao2024tradingagents} replicates a financial market where specialized agents (Risk Managers, Technical Traders) debate investment decisions. The study reveals that organizational diversity leads to emergent market phenomena, such as price discovery, which a single agent cannot simulate. On a larger scale, Generative Agents \cite{park2023generative} demonstrated how information diffuses through a population of agents in a simulated town. Recent work like SocioVerse \cite{zhang2025socioverse} expands this to 10,000+ agents to study social norm propagation. These ``World Models,'' including Genie 3 \cite{deepmind2025genie} and PAN\cite{zhang2025pan}, allow agents to simulate the consequences of actions in a physics-aware latent space before executing them in reality, bridging the gap between internal reasoning and external simulation.

Table \ref{tab:mas_frameworks} presents the state-of-the-art multiagent collaboration frameworks, contrasting their structural rigidity and conflict resolution mechanisms.

\section{Environments and Applications}

An agent is defined not only by its internal architecture but also by the environment in which it operates. Grounding the reasoning capabilities of Large Language Models into actionable execution requires distinct interfaces for different domains. We classify these environments into Digital (Web, OS, Enterprise), Embodied (Robotics), and Scientific domains.

\subsection{Digital Agents: The Web, Operating Systems, and Enterprise}

Most modern knowledge work occurs in digital interfaces, making web and desktop automation a central frontier. The field has moved from passive retrieval toward active execution in open-ended environments.

\subsubsection{The Web Agent Evolution}
Web agency is challenging primarily because interfaces vary widely and evolve dynamically. Mind2Web and WebArena established realistic long-horizon benchmarks across functional site clones (e.g., GitLab, e-commerce) \cite{deng2023mind2web,zhou2023webarena}. Purely text-based agents often break on dynamic DOM changes or canvas elements, motivating multimodal approaches such as WebVoyager, which uses screenshots to infer visual UI structure \cite{he2024webvoyager}. However, visual agents introduce new failure modes: Environmental Distractions shows that models can over-attend to irrelevant UI elements (ads, pop-ups), causing action errors and motivating robustness methods such as task-aware cropping and attention masking \cite{ma2025distractions}.

\paragraph{Native computer use beyond DOM parsing}
Computer-use interfaces push this further by treating the desktop/browser as pixels plus low-level input actions. Anthropic’s computer-use tooling for Claude exposes cursor movement, clicking, typing, and reading screen state \cite{anthropic2024computeruse,anthropic2024computerusetool}, reducing dependence on brittle DOM extraction and app-specific wrappers. OpenAI’s Operator similarly targets GUI operation from screenshots and mouse/keyboard actions \cite{openai2025operator}. These capabilities broaden reachable tasks but expand the attack surface for indirect prompt injection and increase the need for sandboxing, permissioning, and human confirmation on sensitive steps.

\subsubsection{From Browser to Operating System}
Operating-system agents must manage files, switch applications, and recover from partial failures across multi-app workflows. OSWorld benchmarks such end-to-end desktop control and originally highlighted a large gap to human performance due to brittle grounding and long-horizon error accumulation \cite{xie2025osworld}. OSWorld Verified re-evaluates the benchmark to reduce labeling noise and better separate grounding vs.\ planning errors \cite{xlang2025osworldverified}, yielding substantially higher reported performance for top systems; for example, CoAct~1 reports 60.76\% success by combining computer-use actions with coding-as-action for verification and recovery \cite{xin2025coact1}. Windows Agent Arena complements this line by focusing on Windows-specific hierarchies and long-horizon reliability \cite{wu2025windows}, reinforcing that progress depends on both base models and surrounding controllers (grounding, verification, recovery) and should be reported with failure categories (unsafe actions, retries, grounding mistakes), not only mean success.

\subsubsection{Enterprise and Software Engineering}
Enterprise settings emphasize reliability, governance, and extreme horizon lengths. SWE-Bench Pro extends software-agent evaluation toward repository-scale tasks that require hours of simulated work and exposes bottlenecks such as context exhaustion and search-space explosion \cite{deng2025swebenchpro}. For business intelligence, SQL-agent work demonstrates decomposition of complex questions into multi-stage query plans \cite{chen2025sql,chernyshevich2025multihop}. However, deployment requires more than academic accuracy: enterprise frameworks emphasize auditability (trace logs), data governance, and failure recovery—dimensions often absent from general benchmarks such as AgentBench \cite{wang2025enterprise,liu2024agentbench}.

\subsection{Embodied Agents: Robotics and Open-Ended Games}

Embodied AI represents the challenge of grounding language in physical reality, where actions have irreversible consequences and physics dictates constraints. The field has moved from simple instruction following to Vision-Language-Action (VLA) models that directly map perception to motor control.

\subsubsection{Open-Ended Learning in Games}
Before tackling physical robots, agents demonstrated complex behaviors in open-ended game environments. Voyager \cite{wang2024voyager} is the flagship example of an agent capable of lifelong learning. Deployed in Minecraft, Voyager writes its own executable Python code to master novel skills (e.g., combat, mining), stores them in a persistent ``Skill Library,'' and retrieves them to compose complex behaviors. Unlike RL agents that require millions of samples, Voyager explores the world via an automatic curriculum, advancing through the game's technology tree $15.3\times$ faster than baselines. This ``Code-as-Policy'' approach \cite{lynch2023code}  demonstrated that LLMs could reason about long-horizon physical tasks using the abstraction of programming.

\subsubsection{Physical Grounding and Affordances}
Transferring this reasoning to real robots requires addressing ``grounding''—ensuring the agent understands what is physically possible. Early approaches like SayCan \cite{brohan2022saycan} used LLMs as high-level planners, filtering semantic plans through a learned value function of robot affordances (i.e., ``can I actually pick up this cup?''). This paradigm has evolved into end-to-end VLA models. VLA Architectures \cite{din2025vla} catalog over 100 recent models, defining a new taxonomy based on ``Multimodal Alignment''—how vision encoders and language tokens are fused directly into action tokens.

\subsubsection{VLA Models}
The current frontier is represented by systems like Gemini Robotics 1.5 \cite{deepmind2025gemini}. Unlike Voyager's code generation, Gemini 1.5 introduces ``native thinking'' for robots, allowing them to process video input and reason internally about a task (e.g., ``The cup is fragile, I must grip gently'') before generating motor commands. Crucially, it leverages cross-embodiment learning, transferring skills learned on one robot morphology to another, a key step toward general-purpose robotic brains. To further enhance robustness, VLM-GroNav \cite{elnoor2024vlm} integrates proprioceptive sensing (force/torque feedback) with vision. By grounding VLM outputs in physical feedback, robots can detect hazards (e.g., slippery terrain) that vision alone misses, improving navigation success by 50\%.

\subsubsection{Autonomous Driving}
In the high-stakes domain of Autonomous Driving, agents must reason about rules and social intent. Agent-Driver \cite{mao2024agentdriver} departs from traditional black-box neural networks by introducing an explicit reasoning engine. The agent uses a ``cognitive memory'' of traffic rules and a ``Chain-of-Thought'' planner to explain its decisions (e.g., ``Yielding because the pedestrian is entering the crosswalk''), improving safety and interpretability. However, running these massive models on vehicles is computationally prohibitive. To address this latency bottleneck, DiMA \cite{hegde2025dima} proposes a knowledge distillation framework. DiMA compresses giant multimodal models (like GPT-4V) into compact, edge-deployable models, preserving the reasoning logic while reducing parameters by $100\times$, essential for real-time safety.

\subsection{Specialized Domains}

While generalist agents attract attention, many near-term impacts are emerging in specialized verticals where agents must integrate with rigorous workflows and meet safety, regulatory, and traceability requirements.

\subsubsection{Healthcare and Scientific Discovery}

In the natural sciences, agents are shifting from reference tools to research partners. Scientific Intelligence outlines scientific-agent roles spanning hypothesis generation, experiment design, and analysis \cite{ren2025scientific}, and full-loop systems demonstrate end-to-end cycles in which agents propose hypotheses, write Python for simulations, and interpret results with minimal human intervention \cite{zhou2025scientific}. Surveys also report applications in compound and nanobody discovery, while platforms such as ToolUniverse coordinate access to scientific tools and databases via manager/composer architectures \cite{so2025scientific,kempner2025tooluniverse}.

In clinical settings, the emphasis is precision, patient safety, and legal accountability. Medical agents increasingly connect to EHRs, reason over longitudinal histories, and support clinical decision-making and administrative workflows \cite{wang2025medicine,nature2025healthcare}. Agentic Healthcare and MindGuard describe tool-chaining for genomics and imaging and mobile-sensor monitoring for proactive support under audit trails for governance and liability \cite{yuan2025agentic,acm2025mindguard}. Correspondingly, evaluation work argues that generic NLP benchmarks are insufficient and calls for medical-specific tests that combine notes with images and verify compliance with professional guidelines \cite{chen2025evaluation}.

\subsubsection{Finance, Economics, and Advanced Conversational Agents}

In finance, agentic AI supports both automation and market simulation. PyMarketSim offers a controlled environment where RL and LLM agents trade in realistic order books \cite{mascioli2024pymarketsim}, and financial-reasoning studies suggest heterogeneous agent behaviors can reproduce phenomena such as liquidity provision, microstructure exploitation, and bubble-like dynamics, motivating agent-based simulation as a tool for macroeconomic research \cite{lopez2025financial}.

Conversational agents optimize for engagement and social interaction rather than profit. Proactive conversational AI spans reactive systems through deliberative agents that introduce topics and pursue long-term objectives (e.g., education, sales) \cite{acm2025proactive}. Empathetic voice agents combine speech generation with affect modeling so tone and pacing adapt to users, and long-term support agents aim to reduce loneliness via sustained, reliable interactions \cite{ijisae2024empathetic,alotaibi2024support}. Because real-time dialogue is latency sensitive (even sub-second delays degrade presence), deployments emphasize aggressive optimization such as distillation and speculative decoding \cite{rjpn2024realtime}.

\section{Evaluation and Safety}

As agents transition from closed sandboxes to real world deployment evaluation methodologies must evolve beyond simple text similarity metrics like BLEU or ROUGE. Integrating performance metrics \cite{khamis2025agentic} with enterprise deployment requirements \cite{wang2025enterprise}, we adopt the CLASSic framework \cite{wornow2025classic} to assess the field in  five critical dimensions including Cost, Latency, Accuracy, Security, and Stability.

\subsection{Cost: The Efficiency Trade Off}
\begin{figure}[ht]
\centering
\includegraphics[width=\columnwidth]{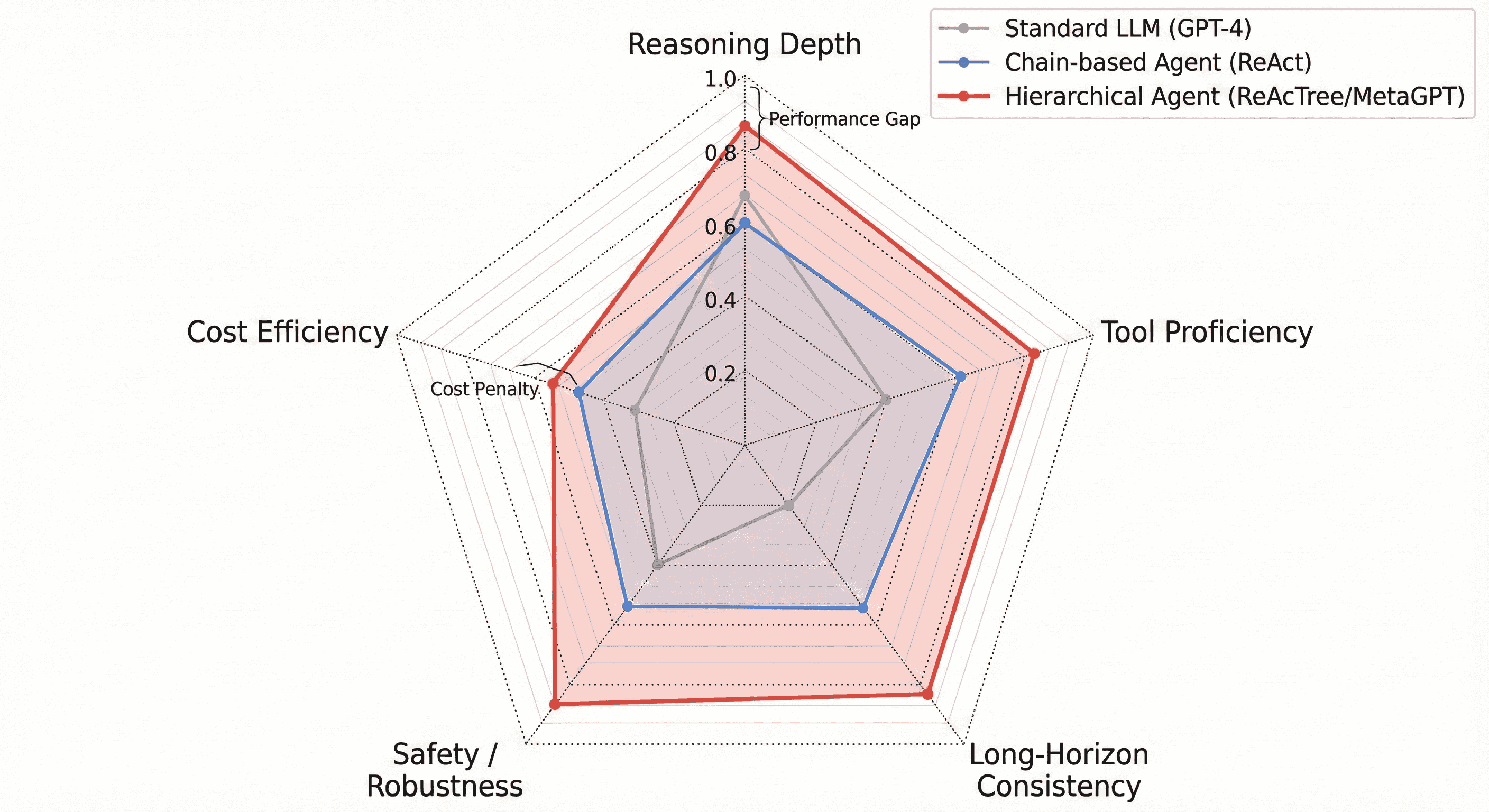}
\caption{\textbf{Multidimensional Architectural Comparison.} We compare architectures across the CLASSic dimensions. While Hierarchical Agents (Red) achieve superior Reasoning Depth and Tool Proficiency, they incur a significant Cost Penalty and Latency compared to standard LLMs (Grey).}
\label{fig:radar}
\end{figure}

High reasoning depth often comes at the price of significant computational overhead \cite{epoch2025prices}. As illustrated in Fig. \ref{fig:radar} hierarchical architectures such as ReAcTree maximize task proficiency and reasoning depth but they incur exponential increases in token consumption \cite{aicerts2025rebench} compared to standard linear chains or zero shot prompting. This Efficiency Intelligence Trade off remains a critical bottleneck for deploying agents in cost sensitive real time applications.

\subsection{Latency: Real World Constraints}
Real world tasks are rarely instantaneous necessitating rigorous latency evaluation. Asynchronous benchmarks like Robotouille \cite{gonzalez2025robotouille} reveal that agents frequently fail when tasks involve variable temporal delays such as waiting for a cooking process to finish before acting. The study shows that while synchronous agents achieve 47\% success asynchronous settings cause performance to plummet to 11\% highlighting a critical lack of temporal awareness in current planners. To mitigate latency in safety critical domains like autonomous driving where sub 100ms response times are mandatory DiMA \cite{hegde2025dima} proposes knowledge distillation. This approach compresses giant multimodal planners like GPT-4V into compact edge deployable models preserving the reasoning logic while reducing parameter count by 100 times thereby solving the latency constraint without sacrificing safety logic.

\subsection{Accuracy: The capability gap and saturation}

Accuracy for agents is not captured by static question answering alone: success can collapse when tasks require tool use, state tracking, and long-horizon recovery, reflecting both base-model limits and end-to-end architecture choices (memory, orchestration, grounding, and permissions). GAIA highlights this gap for general assistants on human-easy tasks that require multi-step decomposition, tool use, and verification \cite{mialon2023gaia}. For desktop control, OSWorld Verified reduces evaluation noise and exposes stronger performance under more reliable protocols \cite{xlang2025osworldverified}; CoAct~1 exemplifies a modern design that combines computer-use actions with coding-as-action to validate and repair steps during execution \cite{xin2025coact1}. In software engineering, SWE-bench Verified reduces ambiguity and brittle tests \cite{openai2024swebenchverified,swebench2024leaderboard}, while SWE-bench Pro pushes toward repo-scale, longer-horizon tasks where context management and search control become central bottlenecks \cite{deng2025swebenchpro}. For tool use under policies, $\tau$-bench evaluates multi-turn interaction in domains like retail and airlines \cite{taubench2024}. FrontierMath targets hard-ceiling mathematical reasoning with reduced contamination, stressing inference-time reasoning and verifier-guided approaches \cite{epoch2024frontiermath}.

Consequently, modern agent evaluation increasingly reports not only mean success but also compute budgets, run-to-run variance, and failure severity distributions, aligning accuracy with CLASSic trade-offs in realistic trajectories. AgentBench evaluates agents across diverse environments \cite{liu2024agentbench}, while MultiAgentBench (MARBLE) measures emergent multi-agent behaviors such as negotiation efficiency and consensus formation \cite{zhu2025multiagentbench}.

\subsection{Security: The trust gap and prompt injection}

Security is a primary barrier to deploying agentic systems: once an LLM is connected to executable tools (file I/O, code execution, enterprise APIs), prompt injection can override the intended objective and turn the agent into a confused deputy \cite{liu2024formalizing}. Attacks may be direct (user input) or indirect via untrusted content encountered during tool use (web pages, documents, tickets, tool outputs). The risk is amplified in high-bandwidth observation channels such as computer-use settings, where agents interpret screenshots and execute low-level mouse/keyboard actions \cite{anthropic2024computerusetool,openai2025operator}. Standardized connector layers (e.g., MCP) further increase the need for governance at integration boundaries, including allowlists, authentication, and audit logging \cite{anthropic2024mcp,anthropic2024mcpspec}.

Prompt-only defenses are brittle: PromptArmor reduces the likelihood that injected instructions are followed \cite{chen2025promptarmor}, but adaptive attackers can craft indirect injections that bypass static guards \cite{zhan2025adaptive}. As a result, robust security is increasingly a systems problem: layered mitigations such as constrained tool permissions, compartmentalized sandboxes, explicit user confirmation for sensitive actions, and independent policy/audit components that validate plans before execution, plus operational monitoring and intervention to limit the impact of unsafe behavior in production \cite{aws2024intervention}.

\subsection{Stability: Failure Mode Analysis}
Finally Stability refers to the system variance over repeated runs \cite{scale2025variance} and its resilience to minor perturbations. In stochastic systems like LLM agents a simple Success Rate metric often masks critical reliability issues. A framework for enterprise agents \cite{wang2025enterprise} argues that rigorous evaluation must include failure mode analysis quantifying not just how often an agent succeeds but the severity distribution of its failures distinguishing a benign timeout from a catastrophic data leak. This is particularly vital in regulated domains. In healthcare applications \cite{chen2025evaluation} emphasizes that agents must demonstrate high compliance stability ensuring that clinical decisions consistently align with medical guidelines regardless of prompt phrasing or stochastic sampling temperature. Future benchmarks must therefore report standard deviation and worst case failure scenarios alongside mean performance to provide a true picture of agent readiness.

\section{Challenges and Future Directions}

Despite the rapid proliferation of agentic architectures, the field remains in a nascent stage. While agents can perform impressive feats in controlled sandboxes, their deployment in unconstrained real-world environments is hindered by fundamental limitations in reliability, efficiency, and alignment. In this section, we synthesize the critical open challenges and outline promising directions for future research.

\subsection{Hallucination in Action and Error Propagation}
The most pervasive challenge in agentic AI is the ``hallucination in action'' problem. While a factual error in a chatbot response is merely misleading, a hallucinated action—such as calling a non-existent API endpoint or deleting the wrong file—can lead to irreversible system failures. Agents often fail when the retrieval component provides irrelevant context \cite{jiang2023active} the retrieval component provides irrelevant context, causing the planner to generate flawed execution steps. Furthermore, in multi-step reasoning loops like ReAct, a single error in an early step propagates downstream, leading to ``cascading failures.'' Future work must focus on robust error-recovery mechanisms, potentially leveraging techniques like SelfCheckGPT \cite{manakul2023selfcheck} to validate reasoning steps before execution occurs.

\subsection{Infinite Loops and Agent Paralysis}
Autonomous agents frequently suffer from getting stuck in repetitive loops, continuously retrying a failed action without modifying their strategy. Benchmarks like WebArena report low success rates (often <15\%) on long-horizon tasks, partially due to agents failing to recognize when they are caught in a local optimum. While architectures like Reflexion attempt to mitigate this via verbal feedback, agents still struggle with ``giving up'' or asking for human help appropriately. Developing ``meta-cognitive'' modules that allow an agent to assess its own progress and interrupt futile loops is a critical research direction.

\subsection{Latency and Computational Cost}
The transition from single-agent to multi-agent systems has introduced significant computational overhead. Architectures that rely on extensive debate or tree-search, such as ToT \cite{yao2024tree}, require multiple LLM inference calls for a single user query. This latency is unacceptable for real-time applications. There is a pressing need for 'System 2' thinking to be distilled into efficient 'System 1' reflexes \cite{li2024personal}. Research into ReWOO \cite{xu2023rewoo} offers a path forward by separating planning from execution, but further optimization is required to make agentic AI economically viable at scale.

\subsection{Human-Agent Alignment and Social Norms}
As agents become more autonomous, ensuring they adhere to human values and social norms becomes paramount. An agent optimized solely for task completion might behave ruthlessly—for example, spamming a user's contact list to achieve a ``networking'' goal. Recent work on socially aligned agents \cite{gabriel2024socially} argues that agents must be trained not just on task success, but on adherence to social contracts and safety constraints. This involves moving beyond simple Reinforcement Learning from Human Feedback (RLHF) toward constitutional AI frameworks where agents intrinsically respect boundaries without needing explicit instructions for every edge case.

\subsection{Towards Open-Ended Learning}
Current agents are largely static; they do not evolve their core competencies after deployment. A major frontier is the development of agents capable of open-ended self-improvement. OMNI \cite{zha2023omni} proposes systems that generate their own curricula, seeking out novel tasks to expand their skill capabilities. This aligns with the vision of Voyager \cite{wang2024voyager}, suggesting a future where agents operate as lifelong learners, continuously acquiring, refining, and sharing new skills without human intervention.

\subsection{Theoretical Limits and Optimization} 
Despite empirical success, the theoretical boundaries of agentic AI remain understudied. OMNI \cite{zha2023omni} suggests that for agents to be truly autonomous, they must possess an intrinsic motivation function to generate their own curriculum, a feature currently absent in standard LLMs. Furthermore, optimization remains a bottleneck; TALM \cite{parisi2022talm} and work on active retrieval \cite{jiang2023active} highlight the latency costs of iterative retrieval. Future architectures may need to integrate introspective capabilities \cite{chen2023introspective} to balance the trade-off between expensive external tool calls and cheaper internal introspection.

\section{Conclusions}

Our investigation on Agentic AI landscape suggests that the central design question is shifting from how to prompt a model to how to program and control a complete agent system. This inversion appears most clearly along three dimensions:

\begin{enumerate}
    \item \textbf{Reasoning:} Architectures have moved from myopic single loop solvers such as ReAct to hierarchical and search based systems, and increasingly to reasoning models that internalize parts of search and backtracking at inference time under a controllable compute budget.
    \item \textbf{Action:} The interaction paradigm has expanded from constrained API calls to code as action and computer use actions, enabling agents to operate both structured tool APIs and arbitrary graphical interfaces, with verification and recovery becoming first class design requirements.
    \item \textbf{Collaboration:} Multi agent systems are moving away from unstructured chat loops toward controllable workflow graphs and explicit handoff patterns, improving observability, debuggability, and safety through flow engineering.
\end{enumerate}

At the same time, the transition from chatbot to reliable agent is incomplete. Recent verified evaluations show rapid gains in desktop and operating system control, but they also make clear that success depends strongly on the surrounding controller, including grounding, tool permissions, and recovery loops \cite{xlang2025osworldverified,xin2025coact1}. Closing the remaining gap to human level reliability will require progress on two coupled fronts: i) efficient reasoning and verification that improves robustness without prohibitive cost or latency, and ii) stronger security and governance that can withstand indirect prompt injection and other confused deputy failures when agents are connected to real tools and data.

Ultimately, progress in Agentic AI is unlikely to come from model scale alone. It will depend on architectures that integrate perception, memory, tools, and collaboration in a way that is not only powerful, but also controllable, auditable, and aligned with the constraints of real world deployment.

\end{document}